\documentclass[12pt]{article}
\usepackage[margin=1in]{geometry}
\usepackage{hyperref}
\usepackage{amsmath}
\usepackage{amssymb}
\usepackage{graphicx}
\usepackage{rotating}
\usepackage{makecell}
\usepackage{dcolumn}
\usepackage{booktabs}
\usepackage{setspace}
\usepackage[numbers]{natbib}

\begin{document}
\onehalfspacing

\title{How Understanding Forecast Uncertainty Resolves the Explainability
  Problem in Machine Learning Models}
\author{Joseph L. Breeden\\
Deep Future Analytics LLC\\
breeden@deepfutureanalytics.com}
\date{\today}
\maketitle


\begin{abstract}
When a machine learning model cannot produce a reliable forecast at a given
point in the feature space, no explanation of that forecast is meaningful.
This paper argues that forecast uncertainty is a necessary condition for
explanation validity: high explanatory instability and high forecast
uncertainty share a common cause, so the appropriate response is not to seek more stable explanation
methods but to recognise that no usable forecast exists. 

This paper formalises this as the primacy of forecast uncertainty over
explanatory stability and derives the tools needed to test it out-of-sample (OOS). For
locally linear models (decision trees, gradient boosted trees, and
piecewise linear regression) a log-linear relationship is established
between OOS Hessian instability and OOS local linear RMSE, demonstrating that
explanatory instability and forecast uncertainty are two measurements of the
same underlying generalisation failure. For smooth nonlinear models such as
sigmoid neural networks, out-of-sample Shapley values
($\boldsymbol{\phi}^{\mathrm{OOS}}$) are introduced, replacing the standard
model-prediction-based coalition values with empirical conditional
expectations of outcomes from a held-out calibration set. It is shown that
$\boldsymbol{\phi}^{\mathrm{OOS}} - \boldsymbol{\phi}^{\mathrm{IS}}$
partitions the conformal prediction error across input features: conformal
prediction gives a single scalar uncertainty bound at each point, and the
OOS--IS Shapley difference vector decomposes that same error by dimension,
showing how much each feature contributes to the model's local failure. A
controlled convergence experiment establishes the local linear framework as
a special case of the conformal framework in the limit of a globally linear
model.

Simulation results across three data-generating functions and three model
types confirm that forecast uncertainty subsumes explanatory instability: in
every well-specified experiment, regions of high uncertainty are regions of
high instability, and a reliable forecast is a precondition for a stable explanation.
\end{abstract}

\subsection*{Practitioner Summary}

Machine learning models are increasingly used for high-stakes decisions such
as loan approvals, yet their adoption in regulated industries is slowed by
concerns that their predictions cannot be adequately explained. This paper
shows that the explanation problem is actually a forecast reliability problem.
When a model cannot produce a trustworthy prediction at a given point,
no explanation of that prediction is meaningful, regardless of which
explanation method is used.

The practical recommendation is straightforward: before reporting any
explanation, first check whether the model's forecast is reliable at that
point. Conformal prediction, a well-established statistical technique,
provides a pointwise uncertainty score that can serve as the gatekeeper.
If uncertainty is below a business-defined threshold, issue the forecast
and report standard feature attributions. If uncertainty is too high,
withhold the forecast and invoke a simpler fallback model instead.

This paper also introduces out-of-sample Shapley values, which decompose
the model's local error by input feature. These identify which variables
are responsible for unreliability in a given region, guiding data
collection and model improvement. Importantly, piecewise linear models
such as gradient boosted trees, often adopted for their supposed
transparency, exhibit the same instability near their segment boundaries
as explicitly nonlinear models. The common assumption that tree-based
models are inherently more explainable is not supported.

For compliance teams, model validators, and risk managers, the paper
provides a decision framework that prioritises forecast reliability over
explanation stability, aligning model governance with statistical rigour.

\section{Introduction}

The explainability problem in machine learning has become increasingly
important as complex algorithms are deployed in critical decision-making
applications. In loan underwriting, regulatory concerns about explainability
have been a direct drag on adoption of nonlinear models. Even though the
academic literature on credit risk modelling is extensive
\cite{breeden2021survey},  deployments of machine learning models continue 
to be debated. Traditional approaches to model interpretability rely
on local linear models built from nearby data points
\cite{ribeiro2016should}. This
approach becomes unstable near high-curvature regions and discontinuities,
leading to a general criticism that machine learning models are inherently
unexplainable \cite{lipton2018mythos}, or that interpretable models should
be used in their place \cite{caruana2015intelligible,nori2019interpretml,rudin2019stop}.

In the quest for explainability to meet regulatory requirements, machine
learning deployments rarely consider forecast uncertainty. Unlike the linear
models with which business teams are accustomed, nonlinear models can have
dramatically varying forecast uncertainty across the feature space. The power
of machine learning is in finding pockets of predictability, which implies
that regions of low predictability also exist.

This paper shows that machine learning models do not have an explainability
problem. They have a user problem. A general correspondence exists
between explanatory stability and forecast uncertainty. Weak forecasts
have unreliable explanations. When a forecast is generated, the first
test should be of the forecast uncertainty, not of the explanation. If the
uncertainty is too high, no explanation is needed, because the forecast cannot
be used.

Both forecast uncertainty and explanatory instability are caused
in part by high-curvature regions of the objective function. Looking at the forecast uncertainty in a local linear approximation explains why tools such as LIME \cite{ribeiro2016should} can produce unstable explanations in certain regions. This does not reflect a deficiency in LIME, but rather a region where no reliable forecast is available.

Nonlinear models are not as unexplainable as was thought. Conversely, piecewise
linear models are not as explainable as commonly assumed. ReLU networks and regression trees are often presented as
globally interpretable because each segment has an unambiguous local linear
model \cite{sudjianto2020unwrapping}. However, this explainability is illusory. At segment boundaries,
natural input perturbations cause boundary crossings that exhibit the same
explanatory instability and forecast uncertainty as more nonlinear models. 
Moving from the center to the boundary of a segment sees increasing uncertainty 
and degrading explainability, just as with an explicitly nonlinear model.
In fact, a well-tuned piecewise linear model will place the segment boundaries precisely in high curvature, high uncertainty regions.

This paper makes several contributions. First, and most significantly, the primacy of forecast uncertainty is formalised and tested using distributional tests on conditional instability. Second, this paper introduces out-of-sample Shapley values for smooth nonlinear models: by replacing the standard model-based coalition values with empirical conditional expectations of outcomes from a held-out calibration set, the OOS--IS Shapley difference vector partitions the conformal prediction error across input features. Conformal prediction gives a single scalar uncertainty bound; OOS Shapley decomposes that same error by dimension, identifying which features are responsible for the model's local failure. This is a model development and monitoring tool, not a replacement for IS Shapley in production reporting. Third, all primary metrics are computed out-of-sample (OOS), using held-out calibration observations that were not used to fit the model; a theoretical derivation establishes that OOS Hessian instability is log-linearly related to OOS local linear RMSE, proving that the two are measurements of the same underlying uncertainty rather than independent quantities. Fourth, a controlled convergence experiment establishes the precise conditions under which the local linear and conformal frameworks agree, providing a quantitative criterion for framework selection.

\section{Background}

Forecast uncertainty and explanatory instability have separately been studied
extensively. The present work brings these concepts together with a
consistent framework rather than reinventing the underlying
methods.

\subsection{Forecast Uncertainty Estimation Methods}

Quantifying the forecast uncertainty of machine learning models is crucial
for safe deployment in risk-sensitive fields. The uncertainty literature
distinguishes two sources. \emph{Aleatoric uncertainty} arises from
irreducible noise in the data-generating process: even with a perfect model,
outcomes vary stochastically. \emph{Epistemic uncertainty} arises from
insufficient knowledge, whether due to limited training data, model
misspecification, or extrapolation beyond the support of the training
distribution \cite{gawlikowski2023,pernot2023,hullermeier2021aleatoric}.
The distinction matters for model improvement: epistemic uncertainty can in
principle be reduced with more data or better specification, while aleatoric
uncertainty sets a floor on achievable precision. Both contribute to forecast
error.

With nonlinear models, forecast uncertainty can vary
 across the feature space. Unlike linear models, where forecast uncertainty increases smoothly toward model boundaries, a nonlinear model can have abrupt transitions between regions of high and low uncertainty. The power of machine learning lies in
finding pockets of predictability, which necessarily implies that regions of
low predictability also exist. This spatial heterogeneity means that a single
global uncertainty estimate is insufficient; pointwise or local uncertainty
quantification is required \cite{lakshminarayanan2017simple,vovk2005}.

Several families of methods have been developed to address this problem.
Bayesian approaches place probabilistic priors on model weights and produce
a distribution over predictions rather than a point estimate
\cite{bishop1997bayesian,neal2012bayesian,goan2020bayesian}; Monte Carlo
dropout approximates this Bayesian model averaging by retaining dropout at
inference time \cite{gal2016dropout}. Where full Bayesian treatment is
impractical, deep ensembles estimate predictive uncertainty from the spread of
independently trained models \cite{lakshminarayanan2017simple}. Quantile
regression models specified quantiles of
the response directly without distributional assumptions
\cite{koenker1978regression}. Each of these methods produces a pointwise
uncertainty estimate, but they differ in computational cost, calibration
quality, and the degree to which they separate epistemic from aleatoric
components.

A second approach fits a weighted local linear model to the model's
prediction surface in a neighbourhood of the query point and measures
uncertainty by the weighted RMSE of that fit. When this RMSE is high, the
model is operating in a region of high curvature or discontinuity where a
linear approximation is inaccurate. This local linear RMSE is an IS metric
by construction: the surrogate is fitted and evaluated on training data.
Section~3 derives an OOS variant that evaluates the surrogate against
held-out calibration outcomes instead, separating approximation error from
generalisation error.

Conformal prediction is a fundamentally different approach: rather than
approximating the prediction surface, it calibrates uncertainty directly
against held-out residuals, making it OOS by construction and applicable
to any base predictor without modification.
For a fitted model
$\hat{f}$ and a calibration set, the conformal score for each calibration
point is the residual $s_i = |y_i - \hat{f}(x_i)|$. In its standard form,
conformal prediction guarantees marginal coverage: averaged over all test
points, the true outcome falls within the interval at the nominal rate.
However, marginal coverage can mask poor performance in specific regions of
the feature space, which is precisely the failure mode relevant to
explainability. Local or conditional conformal methods address this by
constructing intervals from calibration points near the query point
$\mathbf{x}_0$, weighted by proximity \cite{guan2023localized}.
The result
is a spatially varying uncertainty estimate $q(\mathbf{x}_0)$ that is OOS
by construction: calibration residuals are computed on observations not used
to fit $\hat{f}$. This conditional approach is the preferred uncertainty
measure for smooth nonlinear models in this paper, because it is faithful
at each point in the feature space rather than merely correct on average,
and does not depend on the quality of any local linear approximation.

\subsection{Measures of Explanatory Stability}

Explanatory stability refers to the degree to which an explanation of a
model's prediction changes under small perturbations of the input or the
model itself. In regulated settings such as credit underwriting, unstable
explanations undermine both compliance and trust. If two nearly identical
applicants receive materially different reason codes, the explanation
mechanism cannot support consistent decision-making. Two sources of
instability can be distinguished. Perturbation instability arises
when small changes to the input at a fixed model produce large changes in
the explanation. Model-class instability, formalised as the Rashomon
effect \cite{fisher2019all},
arises when multiple models with similar predictive accuracy produce
divergent explanations for the same input. The present paper focuses on
perturbation instability, since it is most applicable to the single-model contexts common in business.

Three families of post-hoc explanation methods dominate current practice.
Local linear surrogates, exemplified by LIME
\cite{ribeiro2016should}, fit a weighted linear model to the prediction
surface in a neighbourhood of the query point and use the surrogate
coefficients as feature attributions. Shapley-based methods, particularly SHAP \cite{lundberg2017shap},
attribute the prediction to features by computing the average marginal
contribution of each feature across all possible subsets of features,
where the contribution of a subset $S$ is measured by the model's
expected prediction with those features fixed versus marginalised.
This coalition value function is the object this paper replaces in
Section~3 to obtain OOS Shapley values: empirical conditional expectations
of outcomes from held-out data substitute for model predictions, making
the attribution answerable to actual observations rather than to
$\hat{f}$ alone. Shapley and LIME are complementary rather than
hierarchical: LIME approximates $\hat{f}$ locally with a linear surrogate
and uses the surrogate coefficients as attributions; Shapley uses
$\hat{f}$ directly without linear approximation, averaging over all
feature subsets rather than fitting a surrogate. Neither is a special
case of the other.
\emph{Gradient-based methods}, including saliency maps and integrated
gradients \cite{sundararajan2017axiomatic}, use the model's input gradients
or path integrals as attribution signals.

LIME and SHAP are the most widely adopted in regulated financial services,
where their model-agnostic nature and relatively intuitive outputs have
made them the default tools for regulatory explanation. This paper
accordingly focuses its analysis on local linear and Shapley-based
frameworks, developing OOS variants of each in Section~3.

A substantial body of work has documented instability across all three
families. Alvarez-Melis and Jaakkola \cite{alvarezmelis2018robust} showed
that LIME explanations can vary significantly under small input
perturbations, without requiring any specific structural property of the
model or input space. Adebayo
et~al. \cite{adebayo2018sanity} demonstrated that several gradient-based
saliency methods are insensitive to both the model's learned parameters and
the training labels, producing attributions indistinguishable from random
baselines. Slack et~al. \cite{slack2020fooling} showed that LIME and SHAP
can be systematically manipulated to hide biased behaviour from auditors.
These findings have motivated two broad responses: developing more stable
explanation algorithms, and regularising models to produce smoother
prediction surfaces that admit more stable local approximations
\cite{ross2018improving,cisse2017parseval}. 

The standard formalisation of perturbation stability is Lipschitz continuity
of the explanation mapping: an explanation function $\mathcal{E}$ is locally
stable at $\mathbf{x}$ if the ratio of explanation change to input change is
bounded for all perturbations in a neighbourhood
\cite{alvarezmelis2018robust}. This
worst-case measure captures the maximum sensitivity of the explanation to
small input changes. Complementary approaches include fidelity and infidelity
measures, which test whether feature attributions co-vary appropriately with
model output under perturbation
\cite{yeh2019fidelity}. Section~3 builds on these
foundations, deriving OOS variants of both a worst-case Lipschitz measure and
an average-case Hessian instability measure for the local linear framework,
and introducing OOS Shapley instability most suitable for smooth nonlinear models.

\section{Quantifying Forecast Uncertainty and Explanatory Instability}

These two literatures, forecast uncertainty and explanatory stability, have
developed largely in parallel. Uncertainty estimation focuses on whether a
prediction is reliable. Explainability research focuses on whether a
prediction can be decomposed into interpretable feature contributions. Prior work has noted empirical connections between the two, but has not
established formal priority. Slack et~al.\ \cite{slack2021reliable}
showed that uncertainty estimates can be used to filter unreliable SHAP
explanations, demonstrating a practical correlation without claiming a
logical relationship. Thuy and Benoit \cite{thuy2024explainability}
found that explanation instability and forecast uncertainty tend to
co-occur in high-curvature regions, treating the two as parallel symptoms
of the same underlying model behaviour rather than one as a precondition
of the other. None of these
contributions establishes that forecast uncertainty is a necessary
condition for explanation validity, or demonstrates that high IS instability
in a region of low OOS uncertainty is a property of the model's learned
geometry rather than a signal of generalisation failure. The simulation
results in Section~5 provide direct evidence of this: for the Wave-like
generator fitted by a sigmoid neural network, IS instability is highest
precisely where OOS uncertainty is lowest, because the model has correctly
learned the oscillatory structure and its prediction surface is genuinely
curved there. OOS metrics correctly identify these regions as reliable;
IS metrics do not.

The present paper argues that forecast uncertainty is logically
prior to explanatory stability. If the forecast is unreliable, the
explanation is moot regardless of its internal consistency. The next
subsection reviews the explanatory stability side of this divide.

For the purposes of studying the relationships between uncertainty and instability, this paper will examine in depth a specific subset of the developed measures. Local linear approximation is a natural framework for quantifying both
forecast uncertainty and explanatory stability at a query point. Local linear approximation can provide
both a forecast error diagnostic and a stability measure. Conformal prediction and Shapley values can
be employed as nonlinear generalisations. This section
derives the primary metrics in both local linear and conformal frameworks and
demonstrates the in-sample bias that arises when evaluation is restricted to
training data.

\subsection{The In-Sample / Out-of-Sample Distinction}

An in-sample (IS) metric at query point $\mathbf{x}_0$ is a function of the
fitted model $\hat{f}$ and the training data $X_{\mathrm{train}}$ only. An
OOS metric additionally requires a calibration set of held-out observations
$\{(\mathbf{x}_i, y_i)\}$ not used to fit $\hat{f}$.

This distinction is not merely a matter of implementation. An IS metric
measures properties of $\hat{f}$'s prediction surface: how it bends,
how steeply it slopes, how rapidly its local linear approximation rotates.
An OOS metric measures whether $\hat{f}$'s predictions are accurate on new
observations. The distinction is consequential for the comparison between IS and OOS
local linear RMSE. The IS variant measures how well a local linear model
approximates $\hat{f}$ near $\mathbf{x}_0$:
\begin{equation}
  \mathrm{LL}_{\mathrm{IS}}(\mathbf{x}_0)
  = \sqrt{\frac{\sum_k w_k\bigl[\hat{f}(\mathbf{x}_k)
    - \hat{\boldsymbol{\beta}}(\mathbf{x}_0)^\top \mathbf{x}_k\bigr]^2}
    {\sum_k w_k}}
\end{equation}
where the sum is over training neighbours $\mathbf{x}_k$ of $\mathbf{x}_0$
weighted by proximity. In plain terms, this is the average prediction error
of the local linear surrogate relative to the model's own outputs at
nearby training points; it measures how curved or complex the prediction
surface is, not how accurate the model is.

The OOS variant fits the same surrogate to training
neighbours but evaluates it against actual outcomes at calibration neighbours:
\begin{equation}
  \mathrm{LL}_{\mathrm{OOS}}(\mathbf{x}_0)
  = \sqrt{\frac{\sum_i w_i\bigl[y_i
    - \hat{\boldsymbol{\beta}}(\mathbf{x}_0)^\top \mathbf{x}_i\bigr]^2}
    {\sum_i w_i}}
\end{equation}
where $\hat{\boldsymbol{\beta}}(\mathbf{x}_0)$ is estimated from training
neighbours and the sum is over calibration neighbours of $\mathbf{x}_0$.
In plain terms, this is the average prediction error of the local linear
surrogate relative to held-out observed outcomes; it measures how well the
model actually generalises to new data near $\mathbf{x}_0$.
When the model is well-specified, both agree. When the model overfits or is
misspecified, they diverge: $\mathrm{LL}_{\mathrm{IS}}$ may be low where
the model has memorised training structure, while $\mathrm{LL}_{\mathrm{OOS}}$
is high because that structure does not generalise.

In-sample measures of uncertainty and explainability can decouple substantially. A well-specified nonlinear
model may exhibit high IS sensitivity to perturbations in a high-curvature region precisely
because it has correctly learned a nonlinear relationship there, while its
OOS error in that region is low. Common IS metrics would flag such a model as unstable, whereas an OOS metric can demostrate that it has low instability. An accurate nonlinear model that is sensitive to perturbations fails only in its alignment to user expectations of how rapidly the underlying system behavior changes.

\subsection{Local Linear Uncertainty and Stability Metrics}

The local linear approximation at $\mathbf{x}_0$ is obtained by fitting a
weighted linear regression to $\hat{f}$ evaluated at the $k$ nearest
training neighbours, with Gaussian kernel weights:
\begin{equation}
  \hat{\boldsymbol{\beta}}(\mathbf{x}_0)
  = \arg\min_{\boldsymbol{\beta}}
  \sum_{i=1}^{k} w_i \bigl[\hat{f}(\mathbf{z}^{(i)})
    - \boldsymbol{\beta}^\top \mathbf{z}^{(i)}\bigr]^2
\end{equation}
In plain terms, this finds the straight-line (linear) description of the
model's behaviour near $\mathbf{x}_0$ that best fits the model's own
predictions at nearby points, with closer points given more weight.
The resulting coefficient vector $\hat{\boldsymbol{\beta}}$ is the local
explanation: each entry is the rate at which the prediction changes with
the corresponding input feature, locally.

The solution is
$\hat{\boldsymbol{\beta}} = (\mathbf{Z}^\top\mathbf{W}\mathbf{Z})^{-1}
\mathbf{Z}^\top\mathbf{W}\hat{\mathbf{f}}$,
where $\mathbf{Z}$ is the matrix of neighbour coordinates, $\mathbf{W}$ the
diagonal weight matrix, and $\hat{\mathbf{f}}$ the vector of model
predictions at those neighbours \cite{hastie2009elements}.

\paragraph{OOS Lipschitz Stability.}
This paper defines an OOS variant of the Alvarez-Melis and Jaakkola \cite{alvarezmelis2018robust}
Lipschitz measure that restricts all neighbour lookups to the training pool and
computes the maximum rate of explanation change across $M$ random
perturbations:
\begin{equation}
  L_{\mathrm{OOS}}(\mathbf{x}_0)
  = \max_{m=1,\ldots,M}
  \frac{\|\hat{\boldsymbol{\beta}}(\mathbf{x}_0)
    - \hat{\boldsymbol{\beta}}(\mathbf{x}_0 + \boldsymbol{\delta}_m)\|}
  {\|\boldsymbol{\delta}_m\|}
\end{equation}
where each $\hat{\boldsymbol{\beta}}$ uses $k$ training-pool neighbours only.
In plain terms, this is the largest ratio of explanation change to input
change observed across all tested perturbations: it answers the question
``if I move this applicant's inputs slightly, how much can the explanation
vector change, in the worst case?'' A high value signals that the
explanation is sensitive to small input changes in that region.

\paragraph{OOS Hessian Instability.}
A Hessian instability measure can be defined as the average-case complement to the
worst-case Lipschitz. Let $B \in \mathbb{R}^{M \times N}$ be the matrix
of local linear coefficients at $M$ perturbed points, with mean row
$\bar{\boldsymbol{\beta}}$ and centered matrix
$B_c = B - \mathbf{1}\bar{\boldsymbol{\beta}}^\top$. The slope covariance
matrix is
\begin{equation}
  \Sigma_\beta = \frac{1}{M-1} B_c^\top B_c
\end{equation}
and the Hessian instability is $H_{\mathrm{OOS}}(\mathbf{x}_0)
= \log(\operatorname{tr}(\Sigma_\beta) + \varepsilon)$,
where all neighbour lookups use the training pool only.
In plain terms, the trace of the covariance matrix sums the variance of
each coefficient across perturbations, giving a single number that reflects
how much the local linear explanation rotates or stretches when the query
point is nudged in any direction. The logarithm compresses the scale to
make the relationship with forecast uncertainty approximately linear.

\subsection{Shapley Values, OOS Shapley, and the Nonlinear Framework}
\label{sec:shapley}

For smooth nonlinear models, local linear instability is elevated in
high-curvature regions not because the model is unreliable but because the
local linear approximation becomes a poor representation of $\hat{f}$.
The appropriate attribution method for these models uses $\hat{f}$ directly,
without linear approximation.

\paragraph{In-Sample Shapley Values.}
Shapley values \cite{lundberg2017shap} attribute $\hat{f}(\mathbf{x}_0)$
to input features by averaging marginal contributions across all feature
subsets. For feature $j$ with $N$ features total:
\begin{equation}
  \phi^{\mathrm{IS}}_j(\mathbf{x}_0)
  = \sum_{S \not\ni j}
  \frac{|S|!\,(N - |S| - 1)!}{N!}
  \bigl[v^{\mathrm{IS}}(S \cup \{j\}) - v^{\mathrm{IS}}(S)\bigr]
\end{equation}
where the IS coalition value $v^{\mathrm{IS}}(S)$ is the expected model
prediction with features in $S$ fixed to $\mathbf{x}_{0,S}$ and complement
features marginalised over a background distribution:
\begin{equation}
  v^{\mathrm{IS}}(S, \mathbf{x}_0)
  = \mathbb{E}_{\mathbf{x}_{\bar{S}}}\bigl[
    \hat{f}(\mathbf{x}_S = \mathbf{x}_{0,S},\,\mathbf{x}_{\bar{S}})\bigr]
\end{equation}
Intuitively, the Shapley value for feature $j$ answers the question: how
much does knowing feature $j$'s value change the model's prediction, on
average across all possible contexts formed by the other features? The
coalition value $v^{\mathrm{IS}}(S)$ represents the model's expected
output when only the features in $S$ are available and held fixed at
$\mathbf{x}_{0,S}$; the remaining features are integrated out over a
background distribution. Adding feature $j$ to coalition $S$ produces an
increment $v^{\mathrm{IS}}(S \cup \{j\}) - v^{\mathrm{IS}}(S)$. The
weighting factor $|S|!(N - |S| - 1)!/N!$ counts the proportion of
orderings in which $j$ would be added to a coalition of size $|S|$,
ensuring that the sum over all subsets gives a fair attribution. The
result is the unique attribution satisfying the four Shapley axioms:
efficiency (attributions sum to the total prediction), symmetry (features
with equal contributions receive equal attribution), linearity (additive
models are attributed additively), and dummy (a feature that never changes
the prediction receives zero attribution).

By the efficiency axiom, IS Shapley values sum exactly to the prediction:
$\sum_j \phi^{\mathrm{IS}}_j(\mathbf{x}_0) = \hat{f}(\mathbf{x}_0)
- \mathbb{E}[\hat{f}]$. No observed outcomes $y$ appear in this computation.
IS Shapley values therefore constitute a precise, internally consistent
decomposition of $\hat{f}(\mathbf{x}_0)$, regardless of whether that
prediction is accurate.

\paragraph{Out-of-Sample Shapley Values.}
OOS Shapley values are defined by replacing the model-prediction-based
coalition value $v^{\mathrm{IS}}$ with an empirical conditional expectation
of outcomes from the held-out calibration set. For a subset $S$ of features,
the OOS coalition value is the weighted mean of calibration outcomes among
the $k$ calibration points nearest to $\mathbf{x}_0$ in the $S$-feature
subspace:
\begin{equation}
  v^{\mathrm{OOS}}(S, \mathbf{x}_0)
  = \frac{\sum_{i \in \mathcal{N}_S(\mathbf{x}_0)} w_{S,i}\, y_i}
    {\sum_{i \in \mathcal{N}_S(\mathbf{x}_0)} w_{S,i}}
\end{equation}
where $\mathcal{N}_S(\mathbf{x}_0)$ denotes the $k$ calibration points
nearest to $\mathbf{x}_0$ in the subspace of features $S$ only, and
$w_{S,i}$ are Gaussian kernel weights based on the $S$-subspace distance.
The resulting OOS Shapley values $\boldsymbol{\phi}^{\mathrm{OOS}}
(\mathbf{x}_0)$ decompose the locally estimated conditional expectation
of the outcome, $\hat{\mathbb{E}}[y \mid \mathbf{x}_0] - \mathbb{E}[y]$,
estimated entirely from held-out data.

\paragraph{The OOS--IS Shapley Difference as Model Error Attribution.}
Since both IS and OOS Shapley values are computed by the same weighted
averaging of marginal coalition contributions applied to different coalition
value functions, their difference is linear in the difference of those
functions. By the linearity axiom of Shapley values:
\begin{equation}
  \phi^{\mathrm{OOS}}_j(\mathbf{x}_0) - \phi^{\mathrm{IS}}_j(\mathbf{x}_0)
  = \phi_j\bigl[v^{\mathrm{OOS}} - v^{\mathrm{IS}}\bigr](\mathbf{x}_0)
\end{equation}
In the full-feature case, this difference coalition function evaluates to
the local model error $\hat{\mathbb{E}}[y \mid \mathbf{x}_0]
- \hat{f}(\mathbf{x}_0)$, so the OOS--IS Shapley differences sum to:
\begin{equation}
  \sum_{j=1}^N
  \bigl[\phi^{\mathrm{OOS}}_j(\mathbf{x}_0)
    - \phi^{\mathrm{IS}}_j(\mathbf{x}_0)\bigr]
  = \hat{\mathbb{E}}[y \mid \mathbf{x}_0] - \hat{f}(\mathbf{x}_0)
\end{equation}
The individual differences $\phi^{\mathrm{OOS}}_j - \phi^{\mathrm{IS}}_j$
therefore partition the total local model error across input features.
Conformal prediction gives the same error as a single scalar bound
$q(\mathbf{x}_0)$; the OOS--IS Shapley difference vector decomposes that
same quantity by dimension, with each component indicating how much feature
$j$ contributes to the gap between the model's prediction and the data.
The sign is informative: a positive difference means feature $j$ pulls the
data above the model's prediction; a negative difference means it pulls
below. The scalar norm $\|\boldsymbol{\phi}^{\mathrm{OOS}}(\mathbf{x}_0)
- \boldsymbol{\phi}^{\mathrm{IS}}(\mathbf{x}_0)\|$ summarises the total
magnitude of the decomposition and should correlate with $q(\mathbf{x}_0)$
since both measure the same local model error. This correspondence is tested
empirically in Section~\ref{sec:results}.

When $\hat{f}$ is well-specified and generalises accurately at
$\mathbf{x}_0$, the two Shapley vectors agree and the error norm is near
zero. When the model is misspecified or overfit locally, they diverge, and
the divergence is simultaneously a reliability warning and a diagnostic of
which features are driving the model failure.

\paragraph{OOS Shapley Instability.}
The OOS Shapley instability is the conformal-weighted variance of
$\boldsymbol{\phi}^{\mathrm{OOS}}$ vectors across $M$ perturbations of
$\mathbf{x}_0$:
\begin{equation}
  \Sigma_w = \frac{\sum_m s_m
    (\boldsymbol{\phi}^{\mathrm{OOS}}_m - \bar{\boldsymbol{\phi}})
    (\boldsymbol{\phi}^{\mathrm{OOS}}_m - \bar{\boldsymbol{\phi}})^\top}
    {\sum_m s_m}
\end{equation}
where $s_m = q(\mathbf{x}_0 + \boldsymbol{\delta}_m)$ is the conformal
score at the perturbed point, $\boldsymbol{\phi}^{\mathrm{OOS}}_m$ is the
OOS Shapley vector at the perturbed point, and $\bar{\boldsymbol{\phi}}$
is the weighted mean. The OOS Shapley instability is
$S(\mathbf{x}_0) = \log(\operatorname{tr}(\Sigma_w) + \varepsilon)$.
In plain terms, this measures how much the feature attributions produced
by OOS Shapley change when the query point is slightly perturbed, with
perturbations in higher-uncertainty regions given more weight. A high
value indicates that the OOS attribution is itself sensitive to small input
changes, reinforcing the case for not reporting an explanation at that point.

Table~\ref{tab:framework} summarises the paired metrics for each framework.

\begin{table}[htbp]
\centering
\caption{Paired metrics for the local linear and nonlinear frameworks.
  $\boldsymbol{\phi}^{\mathrm{IS}}$ and $\boldsymbol{\phi}^{\mathrm{OOS}}$
  denote in-sample and out-of-sample Shapley vectors respectively.}
\label{tab:framework}
\begin{tabular}{llll}
\toprule
\textbf{Framework} & \textbf{Attribution} & \textbf{Forecast Uncertainty}
  & \textbf{Instability / Error} \\
\midrule
Local linear
  & $\hat{\boldsymbol{\beta}}(\mathbf{x}_0)$
  & LL OOS RMSE
  & OOS Lipschitz; OOS Hessian \\
Nonlinear
  & $\boldsymbol{\phi}^{\mathrm{IS}}(\mathbf{x}_0)$
  & Local conformal error $q(\mathbf{x}_0)$
  & OOS Shapley instability\\  
\bottomrule
\end{tabular}
\end{table}

\subsection{Framework Convergence}

The local linear and conformal frameworks are two parameterisations of the
same family of OOS uncertainty estimates. $\mathrm{LL}_{\mathrm{OOS}}$ is
conformal prediction with a local linear surrogate as the base predictor. 
Standard local conformal uses $\hat{f}$ as the base predictor. Both evaluate
against the same calibration residuals. Their difference decomposes as:
\begin{equation}
  |\mathrm{LL}_{\mathrm{OOS}}(\mathbf{x}_0) - q(\mathbf{x}_0)|
  \leq \underbrace{[\text{linearisation error}]}_{\text{base predictor}}
  + \underbrace{[\text{aggregation difference}]}_{\text{RMSE vs.\ quantile}}
\end{equation}

The linearisation error is controlled by the ratio $r/L$, where $r$ is the
characteristic neighbourhood radius and $L$ is the characteristic size of
a leaf (a terminal node of a piecewise linear or tree-based model, within
which the model is exactly linear). Models of this type are described in
Section~5; the present analysis applies whenever the base predictor is
piecewise linear.
When $r \ll L$, the surrogate fits and evaluates within a single leaf where
$\hat{f}$ is exactly linear, so linearisation error vanishes. As $r/L$
increases, the surrogate straddles leaf boundaries, accumulating error
uncorrelated with generalisation quality.

In the limit $r/L \to 0$ (or equivalently, a single-leaf globally linear
model), the two methods differ only in aggregation. The correlation between
them then approaches a ceiling determined by the difference between weighted
RMSE and weighted quantile as aggregation functions over the same residuals.

\subsection{The Primacy of Forecast Uncertainty}

Let $U(\mathbf{x}_0)$ denote the OOS forecast uncertainty at query point
$\mathbf{x}_0$, $I(\mathbf{x}_0)$ any explanatory instability measure, and
$\tau$ an application-specific threshold determined by regulatory tolerance
or decision stakes.

When $U(\mathbf{x}_0) \geq \tau$, the forecast $\hat{f}(\mathbf{x}_0)$
should not be used in the decision, and $I(\mathbf{x}_0)$ is irrelevant
regardless of its value. When $U(\mathbf{x}_0) < \tau$, the forecast is
reliable and $I(\mathbf{x}_0)$ is informative.

This is not a claim that $U$ and $I$ are always correlated. It is a claim of
logical precedence: $U(\mathbf{x}_0) < \tau$ is a necessary condition for
any explanation of $\hat{f}(\mathbf{x}_0)$ to be meaningful. This implies
a testable empirical claim: in the region where $U(\mathbf{x}_0) < \tau$,
instability measures should be stochastically lower than in the region where
$U(\mathbf{x}_0) \geq \tau$. This is tested in Section~\ref{sec:results}
using a one-sided Mann-Whitney test on the conditional distributions of
instability given uncertainty region.

\section{Empirical Support}
\label{sec:examples}

To evaluate the behaviour of uncertainty and explanatory instability, the
methodology is applied to several synthetic examples chosen to exhibit regions
of high curvature or discontinuity. A machine learning model is simply a
nonlinear mapping of inputs to output, so the concepts of uncertainty and
explainability can be explored without requiring specific applied data sets.

\subsection{Nonlinear Functions for Testing}

Three synthetic functions are used, chosen to represent qualitatively
different prediction surfaces.

\paragraph{Wave-like Function.}
\begin{equation}
y = w_1 \tanh(5x_1) + w_2 e^{-x_2^2}\sin(10x_2)
  + w_3 \sin(3x_3)\cos(2x_3) + w_4 x_4 e^{-0.5x_4^2} + \varepsilon
\end{equation}
This function exhibits a steep transition in the $x_1$ direction and
band-limited oscillation in $x_3$, presenting a challenging fitting problem
for all model types.

\paragraph{Radial Function.}
\begin{equation}
y = \frac{\sin(5\|\mathbf{x}\|)}{1 + 0.5\|\mathbf{x}\|^2} + \varepsilon,
\quad \|\mathbf{x}\| = \sqrt{x_1^2 + \cdots + x_N^2}
\end{equation}
Radially symmetric with high curvature near the origin and smooth decay at
large radii. Tests the ability of models to capture smoothly varying
nonlinearity.

\paragraph{Sigmoid Function.}
\begin{equation}
y = \sum_{j=1}^{N} w_j \Bigl[a_j \cdot \sigma(b_j x_j + c_j)\Bigr]
  + \varepsilon, \quad \sigma(t) = \frac{1}{1+e^{-t}}
\end{equation}
A sum of independent sigmoid transformations producing a separable smooth
function that all model types can fit accurately with appropriate
hyperparameters.

All features are sampled from $\mathcal{N}(0,1)$. Noise has standard
deviation equal to 2\% of the signal standard deviation and is generated
once per function, shared across all model types. Train/calibration splits
(70\%/30\%) are fixed once per function. All three model types therefore
see identical training and calibration observations, making cross-model
comparisons valid.

\subsection{Piecewise Linear Models}

With any segmented modelling approach the local in-segment structure becomes
inaccurate at segment boundaries. These transitions create regions where
the assumption of local linearity fails.

These concerns apply equally to gradient boosted regression trees
\cite{friedman2002stochastic}, ReLU neural networks, and piecewise linear
regression trees. A well-trained piecewise linear model places its
boundaries exactly at the high-curvature regions of the underlying function.
Both the nonlinear model and the piecewise linear model fitted to the same
data will exhibit uncertainty and instability in the same feature space
regions. The purported global explainability of piecewise linear models
is illusory near those boundaries.

\subsection{Model Types}

Three model types are fitted for each generator.
\textit{Gradient boosted trees (XGBoost)} \cite{chen2016xgboost}: a
piecewise ensemble fitted by stochastic gradient boosting \cite{friedman2002stochastic},
subject to the same boundary effects as other tree-based methods and
appropriately evaluated in the local linear framework.
\textit{Sigmoid neural network (nnet)} \cite{ripley1996pattern}: a single
hidden layer network with sigmoid activation, producing a smooth
everywhere-differentiable surface and the appropriate test case for the
nonlinear framework.
\textit{Piecewise linear regression tree (lmtree from partykit)}
\cite{zeileis2008model}: each leaf fits a local linear regression, the
purest test of the local linear framework.
All models are tuned by cross-validation. All primary metrics are computed on
the held-out calibration set using the final tuned model.

\section{Results}
\label{sec:results}

Table~\ref{tab:cv} summarises the nine experiments with cross-validated RMSE.

\begin{table}[htbp]
\centering
\caption{Simulation experiments and model fit.}
\label{tab:cv}
\begin{tabular}{llcc}
\toprule
\textbf{Generator} & \textbf{Model} & \textbf{CV RMSE}
  & \textbf{Character} \\
\midrule
Wave-like & XGBoost         & 0.085 & Oscillatory, moderate curvature \\
Wave-like & Sigmoid NN      & 0.042 & Well-fit, smooth nonlinear \\
Wave-like & Piecewise Lin.  & 0.359 & Poorly fit, high boundary count \\
Radial    & XGBoost         & 0.269 & High curvature near origin \\
Radial    & Sigmoid NN      & 0.302 & Smooth, captures curvature \\
Radial    & Piecewise Lin.  & 0.292 & Boundary-heavy, moderate fit \\
Sigmoid   & XGBoost         & 0.021 & Near-perfect fit \\
Sigmoid   & Sigmoid NN      & 0.009 & Near-perfect fit \\
Sigmoid   & Piecewise Lin.  & 0.021 & Near-perfect fit \\
\bottomrule
\end{tabular}
\end{table}

\paragraph{The IS/OOS distinction.}
IS and OOS local linear RMSE are poorly correlated for most experiments
(Table~\ref{tab:isoos}), with $R^2$ ranging from 0.001 (Radial/Sigmoid~NN)
to 0.224 (Wave/Piecewise~Linear). The Hessian IS versus OOS agreement is
substantially higher ($R^2$ 0.56 to 0.90), because both use training data
for neighbour search and differ only in pool scope. The near-zero $R^2$ for
Radial/Sigmoid~NN is the clearest empirical demonstration that IS metrics
measure model geometry, not generalisation reliability. The
conformal-weighted and unweighted OOS Hessian are nearly identical
($R^2$ 0.95 to 0.997), confirming that the conformal weighting is
theoretically motivated but practically negligible at perturbation scale
$\sigma = 0.05$; the weighted formulation is retained throughout.

\begin{table}[htbp]
\centering
\caption{$R^2$ for IS versus OOS metric agreement across all nine
  experiments. Low values indicate IS and OOS metrics measure different
  quantities; high values indicate agreement. LL: local linear RMSE.
  Hessian (unw/wtd): unweighted and conformal-weighted OOS Hessian.}
\label{tab:isoos}
\setlength{\tabcolsep}{5pt}
\begin{tabular}{llcccc}
\toprule
\textbf{Generator} & \textbf{Model}
  & \textbf{LL IS/OOS}
  & \textbf{Hessian IS/OOS}
  & \textbf{Hessian unw/wtd} \\
\midrule
Wave-like & XGBoost        & 0.189 & 0.738 & 0.980 \\
Wave-like & Sigmoid NN     & 0.187 & 0.731 & 0.979 \\
Wave-like & Piecewise Lin. & 0.224 & 0.895 & 0.993 \\
Radial    & XGBoost        & 0.075 & 0.775 & 0.977 \\
Radial    & Sigmoid NN     & 0.001 & 0.753 & 0.974 \\
Radial    & Piecewise Lin. & 0.051 & 0.893 & 0.990 \\
Sigmoid   & XGBoost        & 0.020 & 0.556 & 0.954 \\
Sigmoid   & Sigmoid NN     & 0.066 & 0.567 & 0.950 \\
Sigmoid   & Piecewise Lin. & 0.050 & 0.629 & 0.960 \\
\bottomrule
\end{tabular}
\end{table}

\begin{figure}[htbp]
\centering
\includegraphics[width=\textwidth]{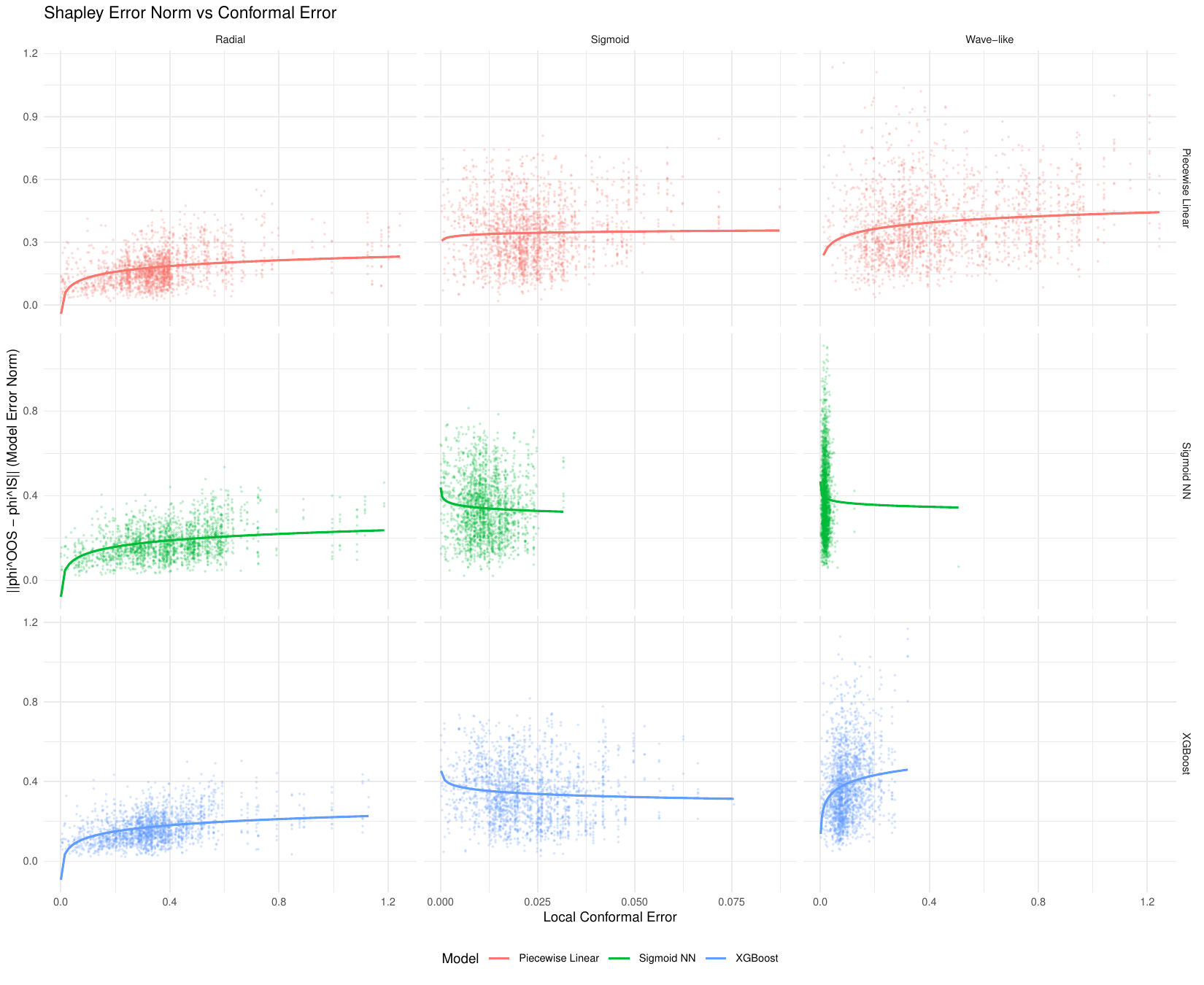}
\caption{OOS--IS Shapley error norm
  $\|\boldsymbol{\phi}^{\mathrm{OOS}} - \boldsymbol{\phi}^{\mathrm{IS}}\|$
  versus local conformal error for all nine experiments (3 generators
  $\times$ 3 model types). Each panel plots one experiment; point colour
  indicates model type. The error norm tracks conformal error positively
  for Radial and Wave-like generators, where conformal error spans a wide
  range, and collapses near zero for the Sigmoid generator where all
  models achieve near-perfect fit.}
\label{fig:errornorm_vs_conformal}
\end{figure}

\begin{figure}[htbp]
\centering
\includegraphics[width=\textwidth]{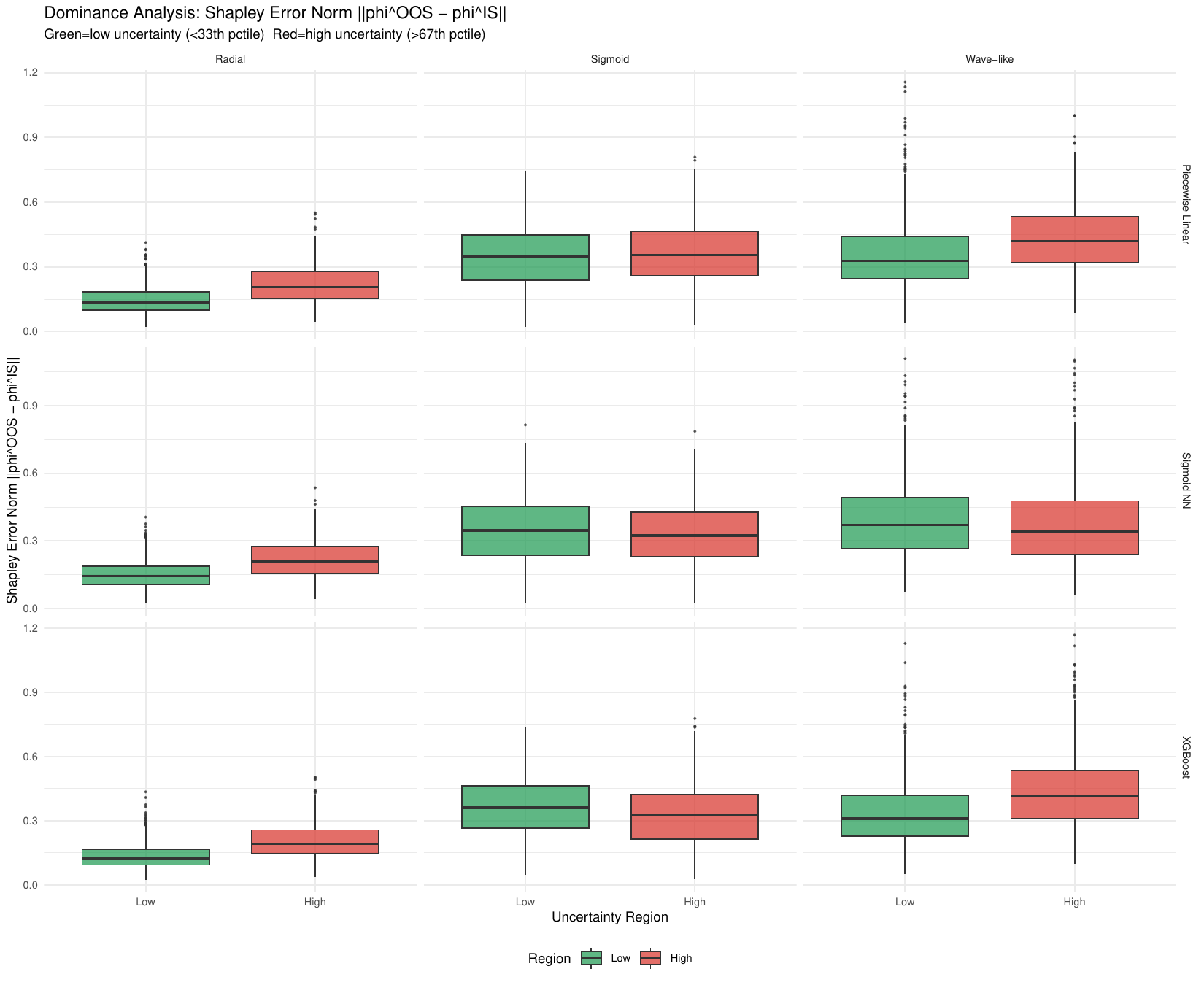}
\caption{Uncertainty primacy test for the OOS--IS Shapley error norm.
  Green boxes: low-uncertainty region (below 33rd percentile of conformal
  scores). Red boxes: high-uncertainty region (above 67th percentile).
  The error norm is stochastically higher in the high-uncertainty region
  for Radial and Wave-like generators across all three model types,
  including Wave-like/Sigmoid~NN where IS instability metrics fail.
  The Sigmoid generator shows negligible separation, consistent with
  near-perfect model fit throughout the feature space.}
\label{fig:uncertainty_errornorm}
\end{figure}

\begin{figure}[htbp]
\centering
\includegraphics[width=\textwidth]{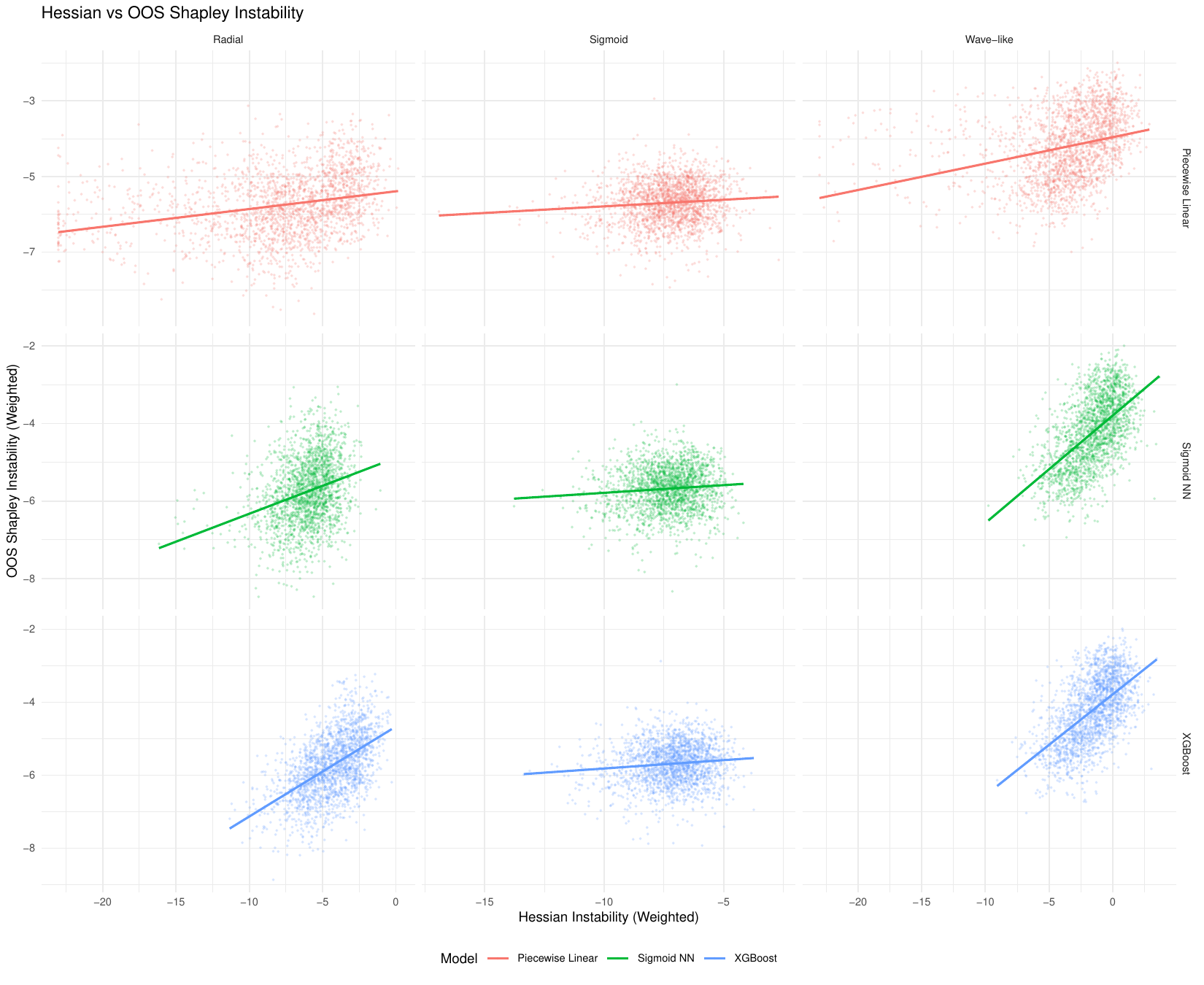}
\caption{OOS Hessian instability versus OOS Shapley instability for all
  nine experiments. The weak correlation confirms that these are non-redundant
  measures: Hessian instability characterises variability in the local linear
  approximation to $\hat{f}$, while OOS Shapley instability characterises
  variability in empirical conditional expectations of $y$ from the
  calibration set.}
\label{fig:hessian_vs_shapley}
\end{figure}

\paragraph{Primacy of forecast uncertainty: local linear framework.}
For the Radial generator the primacy of forecast uncertainty holds across all
model types and instability metrics. Low-uncertainty regions have
stochastically lower instability than high-uncertainty regions, with
Mann-Whitney one-sided $p$-values at or near zero in all cases. For
Radial/Piecewise~Linear, the median LL~OOS~RMSE is 0.172 in the
low-uncertainty region versus 0.375 in the high-uncertainty region; the
median Lipschitz~OOS is 0.325 versus 1.886; the weighted Hessian median
is $-8.26$ versus $-4.92$ (log scale, so smaller is less unstable). For
the Sigmoid generator the primacy of forecast uncertainty holds weakly but in the
correct direction, as all models fit the Sigmoid function well and the
conformal uncertainty range is narrow.

\paragraph{OOS Shapley in the nonlinear framework.}
The OOS Shapley error norm
$\|\boldsymbol{\phi}^{\mathrm{OOS}} - \boldsymbol{\phi}^{\mathrm{IS}}\|$
shows a strong positive relationship with local conformal error across all
nine experiments (Figure~\ref{fig:errornorm_vs_conformal}). The relationship
is tightest for the Radial and Wave-like generators, where conformal error
spans a wide range; the Sigmoid generator collapses to a narrow cluster near
zero as expected, since all model types fit that function accurately and
$\boldsymbol{\phi}^{\mathrm{OOS}} \approx \boldsymbol{\phi}^{\mathrm{IS}}$
throughout. This confirms the theoretical relationship: conformal uncertainty
$q(\mathbf{x}_0)$ is the scalar bound on local model error, and the
OOS--IS Shapley difference vector is the per-feature decomposition of
that same error.

The uncertainty primacy test for the error norm
(Figure~\ref{fig:uncertainty_errornorm}) shows clearly separated
low-uncertainty (green) and high-uncertainty (red) distributions for the
Radial and Wave-like generators across all three model types, with the
high-uncertainty region carrying substantially larger error norms. This
holds even for Wave-like/Sigmoid~NN, where IS instability metrics fail
completely (Mann-Whitney $p \approx 1.0$). The OOS Shapley error norm
therefore correctly identifies the high-uncertainty regions of a
well-fit smooth nonlinear model as genuinely unreliable, a distinction
that IS instability measures cannot make, because they respond to model
curvature rather than generalisation failure.

The Hessian and Lipschitz instability are strongly correlated
across all nine experiments ($r = 0.60$--$0.84$), as expected since both
are linear measures of the same local geometry. The Lipschitz measure estimates the
worst-case rate of explanation change across perturbations, while the
Hessian captures the average-case covariance. 
Within the local linear framework, the Lipschitz measure shows stronger
correlation with LL~OOS~RMSE than the Hessian for piecewise linear models
specifically (Wave/Piecewise, Radial/Piecewise, all three Sigmoid
experiments), whereas the Hessian correlates more strongly for smooth and
ensemble models. For piecewise linear models, the dominant instability
event is a boundary crossing, which is a worst-case rather than an
average-case phenomenon, consistent with the Lipschitz measure being the
more sensitive indicator in that setting.

The OOS Shapley instability shows a similar but weaker separation between uncertainty regions
than the error norm. The Hessian and OOS Shapley instability are weakly
correlated (Figure~\ref{fig:hessian_vs_shapley}), which is the expected
result. Hessian instability is a linear measure, characterising the
covariance of local linear coefficient vectors across perturbations of
$\hat{f}$, while OOS Shapley instability is a nonlinear measure,
characterising variability in empirical conditional expectations of $y$
from the calibration set. The consistently high
Hessian-Lipschitz correlation, combined with the consistently low
Hessian-Shapley and Lipschitz-Shapley correlations, confirms that the two
frameworks characterise different aspects of model behaviour and are not
interchangeable. In practice, the choice of framework is determined by the
model type: the local linear framework (OOS Lipschitz, OOS Hessian) is
appropriate for piecewise linear and tree-based models; the nonlinear
framework (conformal uncertainty, OOS Shapley) is appropriate for smooth
nonlinear models such as sigmoid neural networks. A single deployment
requires only the framework matching its model type. The convergence
experiment shows that when a piecewise linear model approaches global
linearity, the two frameworks agree, so the choice becomes immaterial.

\begin{figure}[htbp]
\centering
\includegraphics[width=\textwidth]{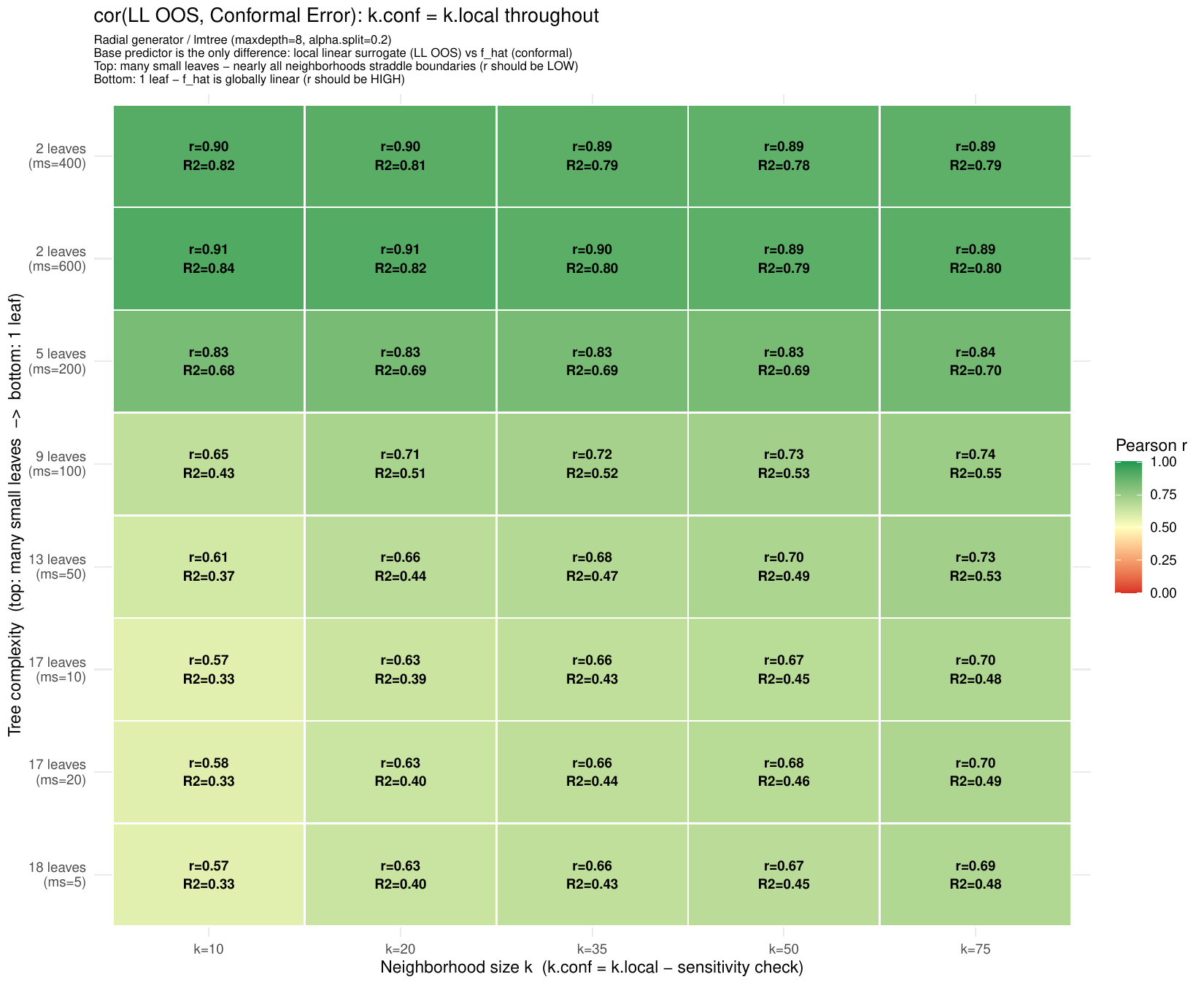}
\caption{Pearson correlation between LL~OOS~RMSE and local conformal
  error as a function of tree complexity (rows, top: many small leaves;
  bottom: 1 leaf) and neighbourhood size $k$ (columns), with
  $k_{\mathrm{conf}} = k_{\mathrm{local}}$ throughout. The correlation
  rises monotonically from $r = 0.57$ at 18 leaves to $r = 0.91$ at 2
  leaves, confirming that the local linear and conformal frameworks
  converge as the model approaches global linearity. The structural break
  between 9 and 5 leaves marks the threshold at which local neighbourhoods
  begin fitting consistently within single leaves.}
\label{fig:v7d_heatmap}
\end{figure}

\begin{figure}[htbp]
\centering
\includegraphics[width=\textwidth]{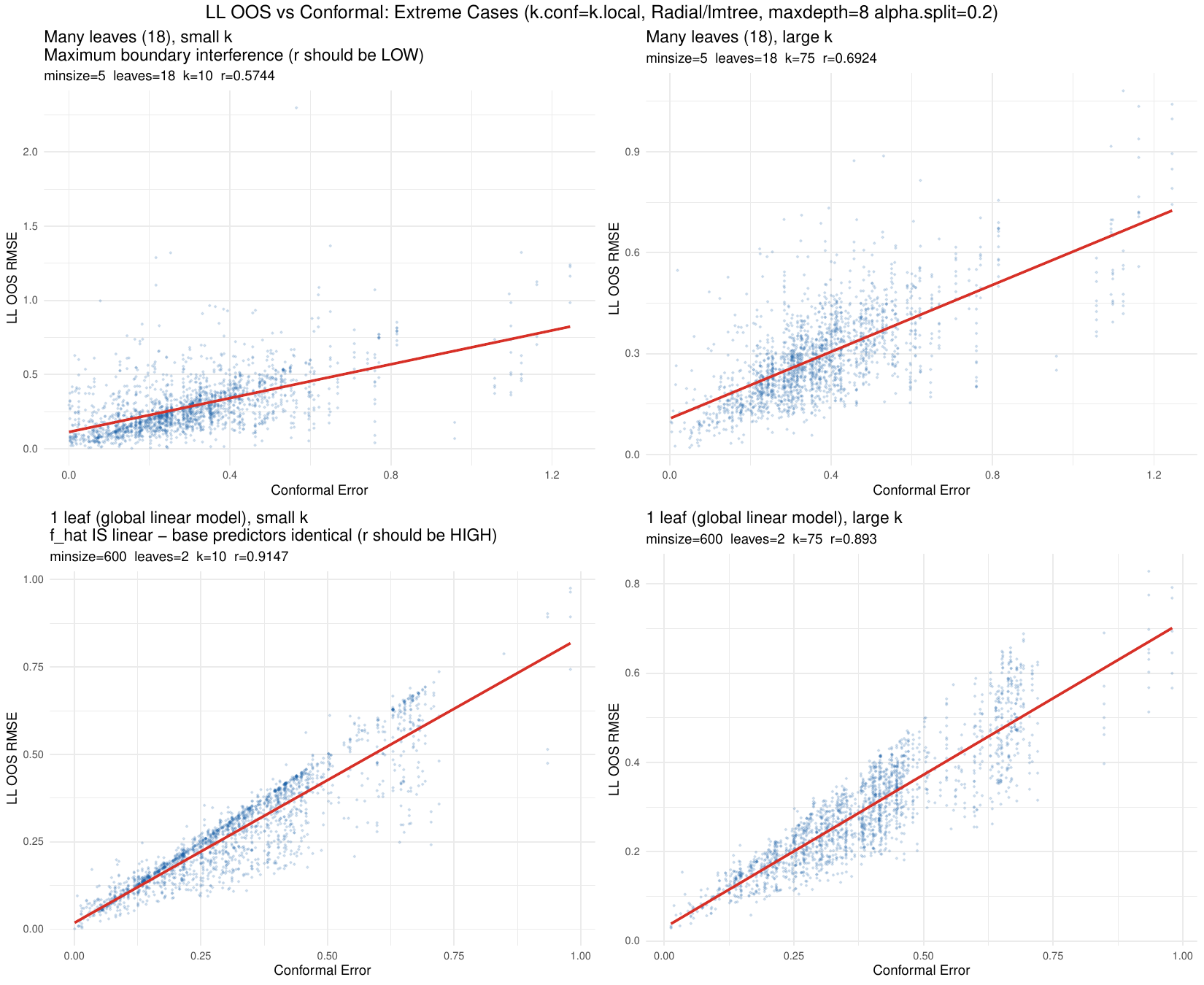}
\caption{LL~OOS~RMSE versus local conformal error at the four extreme
  corners of the convergence experiment. Top left: many leaves ($= 18$),
  small $k = 10$ ($r = 0.57$), maximum boundary interference. Top
  right: many leaves, large $k = 75$ ($r = 0.69$), where noise reduction
  partially offsets boundary straddling. Bottom left: 1 leaf, small
  $k = 10$ ($r = 0.91$), globally linear model with base predictors
  identical. Bottom right: 1 leaf, large $k = 75$ ($r = 0.89$),
  where aggregation noise from a larger neighbourhood slightly reduces correlation.}
\label{fig:v7d_corners}
\end{figure}

\paragraph{When IS instability inverts: the Wave-like generator.}
For the Wave-like generator with Sigmoid~NN and Piecewise~Linear models,
IS instability metrics have Mann-Whitney $p$-values at or near 1.0:
IS instability is higher in the low-uncertainty region than in the
high-uncertainty region. The wave-like function has regions of high local
curvature (peaks and troughs) where a well-fit sigmoid~NN bends steeply.
The local linear approximation is poor there, producing high local linear
instability, but the model is accurate. It has learned the oscillatory
structure correctly and conformal uncertainty is low. In near-zero-curvature
regions the model is locally linear and IS instability is low, but the
noise floor dominates and conformal uncertainty is relatively higher.
This is the expected behaviour of a well-specified smooth nonlinear model,
and it is precisely why the OOS Shapley framework is needed.

\paragraph{Framework convergence.}
The convergence experiment uses the Radial generator with lmtree
(\texttt{maxdepth}$=8$, \texttt{alpha.split}$=0.2$), varying minsize to
produce a tree complexity grid of 18, 17, 17, 13, 9, 5, and 2 leaves.
$k_{\mathrm{conf}} = k_{\mathrm{local}}$ throughout, so that the only
difference between the two methods is the base predictor: a local linear
surrogate for LL~OOS versus $\hat{f}$ for conformal.

The heatmap (Figure~\ref{fig:v7d_heatmap}) shows a clear monotone
increase in the Pearson correlation between LL~OOS and conformal error as
the tree moves from many small leaves toward a single globally linear leaf.
At 18 leaves (the all-boundaries limit, where nearly every local
neighbourhood straddles a leaf boundary) the correlation is $r = 0.57$
($R^2 = 0.33$) at $k = 10$. At 2 leaves the correlation reaches
$r = 0.91$ ($R^2 = 0.84$) at $k = 10$, falling only slightly to
$r = 0.89$ ($R^2 = 0.80$) at $k = 75$.
The residual gap from $r = 1.0$ reflects the difference between weighted
RMSE and weighted quantile as aggregation functions over the same
calibration residuals, not any base-predictor disagreement.

The $k$ axis exhibits two distinct behaviours. At the many-leaves end,
larger $k$ improves the correlation modestly (0.57 to 0.69 for 18 leaves),
because averaging over more calibration points reduces estimation noise
and outweighs the additional boundary straddling that comes with a larger
neighbourhood. At the 2-leaf end, increasing $k$ has almost no effect and
produces a slight decline (0.91 to 0.89 for minsize $= 600$), because
at that point the base-predictor difference is the sole driver and both
methods are averaging over essentially the same linear region regardless
of $k$. The most important finding is the structural break between 9 leaves
($r = 0.65$--$0.74$) and 5 leaves ($r = 0.83$--$0.84$): this is the
threshold at which neighbourhoods begin consistently fitting within single
leaves, and the local linear surrogate becomes an accurate proxy for
$\hat{f}$.

\section{Theoretical Results}

The preceding empirical results revealed some dominant patterns. This section
derives these approximate relationships from first principles.

\subsection{OOS Hessian Instability and OOS Local Linear RMSE}

The log-linear relationship between Hessian instability and local linear
forecast uncertainty, observed in Section~\ref{sec:results},
can be derived from first principles when both quantities are evaluated
out-of-sample.

The OOS local linear RMSE $\mathrm{LL}_{\mathrm{OOS}}(\mathbf{x}_0)$
is defined above. Its square $\sigma^2_{\mathrm{OOS}} \approx
\mathrm{LL}^2_{\mathrm{OOS}}$ captures genuine generalisation error.
It is large where the model fails to predict held-out outcomes and small
where it succeeds.

At each of the $M$ perturbed points $\mathbf{x}_0 + \boldsymbol{\delta}_m$,
treating the calibration residuals as the error process with variance
$\sigma^2_{\mathrm{OOS}}$, the sampling covariance of the local coefficient
vector $\hat{\boldsymbol{\beta}}_m$ is:
\begin{equation}
  \operatorname{Cov}(\hat{\boldsymbol{\beta}}_m)
  = \sigma^2_{\mathrm{OOS}}\,
    (\mathbf{Z}_m^\top \mathbf{W}_m \mathbf{Z}_m)^{-1}
\end{equation}
where $\mathbf{Z}_m$ and $\mathbf{W}_m$ are the neighbour coordinate and
weight matrices at the perturbed point. Taking the trace and averaging
across perturbations:
\begin{equation}
  \operatorname{tr}(\Sigma_\beta)
  \propto \sigma^2_{\mathrm{OOS}}\,
    \operatorname{tr}\!\bigl[(\mathbf{Z}^\top\mathbf{W}\mathbf{Z})^{-1}\bigr]
  = \mathrm{LL}_{\mathrm{OOS}}^2 \cdot C(\mathbf{x}_0)
\end{equation}
where $C(\mathbf{x}_0) = \operatorname{tr}[(\mathbf{Z}^\top\mathbf{W}
\mathbf{Z})^{-1}]$ is a local geometry factor that varies slowly when the
neighbourhood is approximately isotropic. Therefore:
\begin{equation}
  H_{\mathrm{OOS}}(\mathbf{x}_0)
  = \log\bigl(\operatorname{tr}(\Sigma_\beta)\bigr)
  \approx 2\,\log\bigl(\mathrm{LL}_{\mathrm{OOS}}(\mathbf{x}_0)\bigr)
    + \log C(\mathbf{x}_0)
\end{equation}
predicting a log-linear relationship between OOS Hessian instability and
OOS local linear RMSE, with slope approximately 2 and an additive geometry
term. This is exactly the functional form $y = \alpha + \beta\log(x)$
observed in the simulation results, here derived from the OOS error process.
The derivation holds whenever the local linear surrogate is a reasonable
approximation to $\hat{f}$ in the neighbourhood, which is the defining
condition of the local linear framework.

In plain terms, the derivation shows that explanation instability and
forecast uncertainty are not two separate things that happen to correlate:
they are both consequences of the same quantity, the error variance of the
local linear surrogate on held-out data. When the model generalises well
near $\mathbf{x}_0$, that variance is small, the local linear coefficients
are stable, and the explanation is trustworthy. When generalisation fails,
the variance is large, the coefficients are unstable, and the explanation
is not. The log-linear relationship between the two is not an empirical
coincidence but a mathematical identity.

The OOS Lipschitz measure, as a worst-case rather than average-case
quantity, does not admit this precise derivation: it is determined by a
single extremal perturbation rather than by the covariance of the full
distribution. Nevertheless, because it characterises the same local
geometry as the Hessian from a worst-case perspective, it shows an
approximate empirical relationship to OOS local linear RMSE across the
simulation experiments. The relationship is strongest for piecewise linear
models, where boundary crossings are the dominant perturbation event and
the worst-case Lipschitz measure captures those crossings more directly
than the average-case Hessian.

\subsection{The Necessity of Out-of-Sample Shapley Values}

Standard IS Shapley values satisfy the efficiency axiom:
$\sum_j \phi^{\mathrm{IS}}_j(\mathbf{x}_0) = \hat{f}(\mathbf{x}_0)
- \mathbb{E}[\hat{f}]$. If $\hat{f}(\mathbf{x}_0)$ is a poor estimate
of $\mathbb{E}[y \mid \mathbf{x}_0]$, then IS Shapley values are a precise
and internally consistent decomposition of an unreliable number. No property
of Shapley values prevents a model from being stably and coherently wrong.
Low IS Shapley instability and high conformal error can coexist, and in such
cases the IS attribution is misleading regardless of its stability.

OOS Shapley values resolve this by replacing $\hat{f}$ with empirical
conditional expectations of $y$ from the calibration set, as defined in
Section~\ref{sec:shapley}. The key theoretical result, proved using the linearity axiom
of Shapley values, is that the OOS--IS difference vector
$\boldsymbol{\phi}^{\mathrm{OOS}} - \boldsymbol{\phi}^{\mathrm{IS}}$
constitutes a feature-level attribution of the local model error
$\hat{\mathbb{E}}[y \mid \mathbf{x}_0] - \hat{f}(\mathbf{x}_0)$,
with individual components identifying which features are responsible
for the discrepancy between prediction and data.

Conformal uncertainty $q(\mathbf{x}_0)$ and the OOS--IS Shapley difference
vector are therefore two views of the same quantity: $q(\mathbf{x}_0)$ is
the scalar bound on local model error, and
$\boldsymbol{\phi}^{\mathrm{OOS}} - \boldsymbol{\phi}^{\mathrm{IS}}$
is its per-feature decomposition, partitioning that error across input
dimensions with the signs indicating the direction of each feature's
contribution. In production,
conformal uncertainty alone is sufficient to determine whether a forecast
should be issued: when $q(\mathbf{x}_0) < \tau$, the forecast is reliable
and IS Shapley attributions are trustworthy; when $q(\mathbf{x}_0) \geq
\tau$, no forecast is issued and no explanation is needed. OOS Shapley
values are not required in this decision loop. Their value is as a model
development and monitoring tool: the feature-level decomposition of model
error identifies which inputs are responsible for unreliability in a given
region, guiding data collection, feature engineering, or model respecification.

\section{Fallback and Safe Decision-Making at High-Uncertainty Regions}

Rather than focusing on explanatory stability, deployment systems must
incorporate forecast uncertainty into the decision logic. When forecast
uncertainty is too high, the primary concern should be obtaining a usable
decision, not explaining a weak decision.

Nonlinear machine learning models cannot be expected to provide usable
forecasts throughout the feature space. When uncertainty crosses a
predefined acceptance threshold $\tau$, a fallback model should be employed.
This necessarily means that all machine learning models should be evaluated
systematically across the feature space to identify unpredictable regions.

The fallback will most often be a simpler linear model, such as logistic
regression. Such models are unlikely to have predictability gaps, though they
will tend to be less predictive than a machine learning model in regions
where the machine learning model performs well.
The threshold $\tau$ should be set according to acceptable business risk and
may be a function of the forecast value: a large uncertainty at low credit
default risk may be irrelevant, whereas the same uncertainty at high default
risk could be consequential.

In practice, $\tau$ will often need to be asymmetric with respect to the
decision outcome. In lending, for example, regulatory frameworks impose
stricter requirements on adverse actions than on approvals. Denying credit
requires a defensible explanation, which in turn requires a reliable
forecast. A deployment system might therefore apply a tighter threshold when
the forecast would trigger a denial than when it would support an approval.
The production decision workflow is therefore: compute OOS forecast
uncertainty at the query point; if $q(\mathbf{x}_0) < \tau$, issue the
forecast and report IS Shapley attributions; if $q(\mathbf{x}_0) \geq
\tau$, do not issue the forecast, invoke the fallback model, and verify
that the fallback model's own uncertainty at that point is acceptable.
OOS Shapley values play no role in this loop. They are a model development
and monitoring tool: by identifying which features account for the gap
between prediction and data in high-uncertainty regions, the OOS--IS
Shapley error norm informs whether additional data collection, feature
engineering, or model respecification could extend the model's reliable
operating region. The fallback model should be maintained in active
operation, not as a dormant backup, since a model that has not been
monitored against current data cannot be trusted when it is suddenly needed.

\section{Conclusions}

Realizing that machine learning models are powerful because they can identify pockets of predictability, logically implies many of the findings in this paper. A model with high predictability regions consequently has regions of low predictability, and a low certainty forecast should not be used, regardless of whether it can be explained. 
The observed instability of explanations in some regions is shown to be a symptom of the
forecast failure there, not an independent problem to be solved. Previous work has
connected uncertainty and explainability empirically but has not made this
logical priority explicit, nor the implication for decision-making. Making the forecast uncertainty primacy explicit changes the correct sequence of
operations: compute OOS forecast uncertainty first, treat it as a necessary
condition for issuing any forecast, and interrogate explanatory stability
only where that condition is satisfied.

This paper provides the theoretical and empirical machinery to act on that
principle. For locally linear models it is shown, via weighted least squares
covariance propagation, that OOS Hessian instability is log-linearly related
to OOS local linear RMSE, representing two measurements of the same underlying
generalisation failure rather than two independent quantities. For smooth nonlinear
models, out-of-sample Shapley values are introduced, replacing the
model-prediction-based coalition values of standard SHAP with empirical
conditional expectations of outcomes from a held-out calibration set.
The difference $\boldsymbol{\phi}^{\mathrm{OOS}} -
\boldsymbol{\phi}^{\mathrm{IS}}$ is a feature-level attribution of local
model error, identifying not only that the model is wrong at a given point
but which features are responsible, with conformal uncertainty as the scalar
summary of that divergence. A controlled convergence experiment established
that the local linear framework is a special case of the conformal framework,
with the ratio of neighbourhood radius to leaf size as the governing parameter.

The simulation results confirm that in every well-specified case, forecast
uncertainty subsumes explanatory instability. The one apparent failure mode (a well-fit smooth nonlinear model that has
genuinely learned high-curvature structure) resolves on inspection. In-sample
instability metrics flag model
complexity rather than generalisation failure, while OOS metrics correctly
identify those regions as reliable. That distinction is a clear illustration of the principle in action.

For practitioners in regulated industries, the recommendation is direct:
before reporting any explanation, verify that the model produces a reliable
forecast at that point. Conformal uncertainty is the production gatekeeper:
where it is below the acceptance threshold, issue the forecast and report
IS Shapley attributions; where it exceeds the threshold, do not issue the
forecast and no explanation is needed. OOS Shapley values are a development
and monitoring tool, not a production requirement: they identify which
features drive model error in unreliable regions, supporting ongoing model
improvement. Piecewise linear models, often deployed for their purported
transparency, offer no exemption. Near segment boundaries they exhibit the
same forecast uncertainty and explanatory instability as any nonlinear
model, and for the same reason.


\section*{Disclosure of Interest}

The author reports no competing interests.

\section*{Declaration of Funding}

No funding was received for this research.

\bibliography{complete2.bib,ml_explain.bib}
\bibliographystyle{elsarticle-num}

\end{document}